\documentclass{article}

\PassOptionsToPackage{numbers, compress}{natbib}

\usepackage[final]{neurips_2025}

\usepackage[utf8]{inputenc} 
\usepackage[T1]{fontenc}    
\usepackage{hyperref}       
\usepackage{url}            
\usepackage{booktabs}       
\usepackage{amsfonts}       
\usepackage{nicefrac}       
\usepackage{microtype}      
\usepackage{xcolor}         

\usepackage{times}
\usepackage{latexsym}

\usepackage{inconsolata}

\usepackage{amsmath,amsfonts,bm}

\def\eqref#1{equation~\ref{#1}}

\def\1{\bm{1}}

\DeclareMathAlphabet{\mathsfit}{\encodingdefault}{\sfdefault}{m}{sl}
\SetMathAlphabet{\mathsfit}{bold}{\encodingdefault}{\sfdefault}{bx}{n}

\def\gA{{\mathcal{A}}}

\def\gS{{\mathcal{S}}}

\newcommand{\E}[2]{\mathbb{E}_{#1} \left[#2\right]}

\usepackage{hyperref}       
\usepackage{url}            
\usepackage{booktabs}       
\usepackage{amsfonts}       
\usepackage{nicefrac}       
\usepackage{xcolor}         
\usepackage[pdftex]{graphicx}
\usepackage{amsmath}
\usepackage{diagbox}
\usepackage{amsthm}
\usepackage{algorithmic}
\usepackage{algorithm}
\usepackage{bbm}
\usepackage{enumerate}
\usepackage{caption}
\usepackage{wrapfig}
\usepackage{tabularx,ragged2e}
\usepackage{enumitem}

\renewcommand{\hat}{\widehat}

\usepackage{amsmath}
\usepackage{amssymb}
\usepackage{mathtools}
\usepackage{amsthm}

\usepackage[capitalize,noabbrev]{cleveref}

\title{Planning without Search: Refining Frontier LLMs with Offline Goal-Conditioned RL}

\author{%
Joey Hong \quad Anca Dragan \quad Sergey Levine\\
UC Berkeley \\
\texttt{\{joey\_hong,anca,sergey.levine\}@berkeley.edu}
}

\begin{document}

\maketitle

\begin{abstract}
Large language models (LLMs) excel in tasks like question answering and dialogue, but complex tasks requiring interaction, such as negotiation and persuasion, require additional long-horizon reasoning and planning. Reinforcement learning (RL) fine-tuning can enable such planning in principle, but suffers from drawbacks that hinder scalability. In particular, multi-turn RL training incurs high memory and computational costs, which are exacerbated when training LLMs as policies. Furthermore, the largest LLMs do not expose the APIs necessary to be trained in such manner. 
As a result, modern methods to improve the reasoning of LLMs rely on sophisticated prompting mechanisms rather than RL fine-tuning.   
To remedy this, we propose a novel approach that uses goal-conditioned value functions to guide the reasoning of LLM agents, that scales even to large API-based models. 
These value functions predict how a task will unfold given an action, allowing the LLM agent to evaluate multiple possible outcomes, both positive and negative, to plan effectively. 
In addition, these value functions are trained over reasoning steps rather than full actions, to be a concise and light-weight module that facilitates decision-making in multi-turn interactions.
We validate our method on tasks requiring interaction, including tool use, social deduction, and dialogue, demonstrating superior performance over both RL fine-tuning and prompting methods while maintaining efficiency and scalability.
\footnote{Website can be found at \url{https://jxihong.github.io/pnlc_website/}}
\end{abstract}

\section{Introduction}

\begin{wrapfigure}{R}{0.45\linewidth}
\centering
\includegraphics[width=\linewidth]{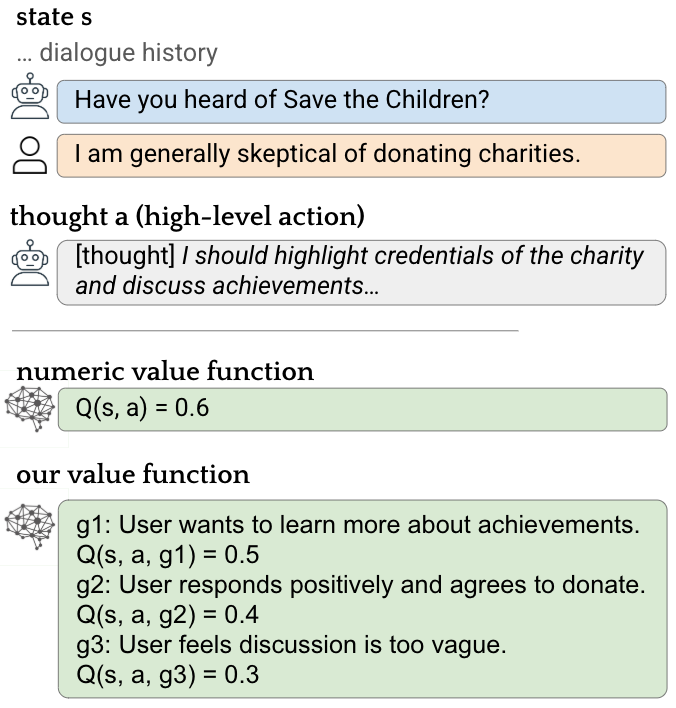}
\caption{In an ongoing goal-oriented dialogue, we learn natural language value over the internal reasoning steps of an LLM agent.
The value analyzes future positive and negative outcomes, and allow the LLM agent to refine its reasoning.
}
\label{fig:dialogue_example}
\vspace{-0.2in}
\end{wrapfigure}
Large language models (LLMs) have demonstrated impressive performance across a wide range of reasoning and problem-solving tasks~\citep{wei2023chainofthought,kojima2022largelanguagemodelszeroshot,yao2023treethoughtsdeliberateproblem}. However, many complex tasks, such as goal-oriented dialogue, social interaction, and negotiation require LLMs to engage in complex multi-turn interactions, where the LLM agent might need to act strategically, anticipating the long-horizon outcomes of its decisions, while still adapting to unexpected responses. In such settings, an effective agent might take actions that are not immediately rewarding, but instead lay the ground for future success by either eliciting information or changing the interlocutor's disposition. For example, an agent may ask clarifying questions to gather crucial information or attempt to persuade an interlocutor to shift their opinion, ultimately enabling the agent to achieve its objective. The space of possible outcomes when planning out such complex behaviors is enormous, and enabling such agents to act intelligently and strategically presents a major algorithmic challenge.

A traditional approach to equipping LLMs agents with such planning capabilities involves fine-tuning them using multi-turn reinforcement learning (RL) \citep{snell2022offline,qtransformer}. While RL can in theory help LLMs optimize their behavior across extended interactions, the approach does not scale well to frontier models like \texttt{GPT-4}~\citep{OpenAI2023GPT4} or \texttt{Claude 3}~\cite{TheC3}. 
In particular, RL algorithms are challenging due to their complexity to train and need for high number of samples, both of which make training large LLMs prohibitively expensive.
Consequently, while RL has achieved some successes in multi-turn settings, its scalability limitations hinder broader adoption.
Because of this, recent methods that leverage frontier models as LLM agents do not perform RL fine-tuning, but during inference, prompt the agent to simulate future trajectories to guide decision-making~\citep{yu2023promptbasedmontecarlotreesearch,zhou2024languageagenttreesearch}.

To address these challenges, we propose a novel approach that leverages offline RL to significantly improve an LLM agent's reasoning and planning, \emph{without directly training the LLM agent}. Our key insight is that at rather than learning a policy, we instead learn a \emph{natural language critic} that can be used by an LLM agent to evaluate potential actions at inference-time. 
Specifically, we train a goal-conditioned value function from offline data that predicts the likelihood of achieving various outcomes given the current state and a proposed action. 
At inference, the critic uses these values to assess future outcomes, which the LLM agent can use to effectively and efficiently guide its planning.

Unlike standard scalar-based value functions, our natural language value functions can provide predictions about a wide variety of potential outcomes, allowing the agent to compare different strategies along many dimensions. 
Figure~\ref{fig:dialogue_example} illustrates this concept in a persuasion task, where the agent evaluates a potential response based on its likely impact.
In the example, the LLM agent learns that by using a credibility appeal, there is $40\%$ probability that the user feels assured and responds positively, but a $30\%$ change the user does not trust the LLM agent for being too vague. These predictions can be processed by the LLM agent to determine which of the available strategies would be most successful.

We further make a number of design decisions that provide for a computationally efficient and scalable method.
First, rather than evaluating low-level utterances, we operate on high-level strategies, or thoughts, akin to chain-of-thought reasoning~\citep{wei2023chainofthought}. 
This abstraction significantly reduces the complexity of the decision space while preserving essential information.
In this hierarchical design, our natural language critic assists the LLM agent in planning over high-level strategies, which can then be used to generate actions in the environment.
By leveraging the critic, our approach also eliminates the need for any inference-time search, significantly improving its practicality on time-sensitive tasks. 
Instead of existing methods rely on feedback from inference-time search, we simply require evaluations by our natural language critic on both positive and negative hypothetical futures --or desirable and undesirable future occurrences to accomplishing the task at-hand, respectively.

Finally, we validate our method across diverse LLM benchmarks requiring interactive decision-making, including web navigation, social deduction games, and persuasion via dialogue. Our approach consistently outperforms state-of-the-art techniques based on fine-tuning with multi-turn RL and inference-time search with self-refinement. Furthermore, it achieves these improvements with significantly lower computational costs, making it a scalable solution for real-world applications.

In summary, our contributions are threefold: (1) we introduce goal-conditioned value functions that model the probabilities of various outcomes, enabling LLMs to reason about long-term effects of their actions; (2) we apply these value functions at the level of high-level thoughts rather than low-level utterances, reducing the complexity of decision-making; and (3) we introduce a novel inference-time process where the LLM agent considers both positive and negative future outcomes to refine its decisions. By integrating these innovations, we provide an effective framework for LLM agents to adaptively reason and plan in multi-turn interactive settings.
Compared to existing state-of-the-art results, our method scales to the largest frontier models with very lightweight training, and does not require a substantial inference budget. 


\section{Related Work}

Our method combines concepts from LLM agents, goal-conditioned reinforcement learning, and test-time prompting, reasoning, and planning methods for LLMs. We summarize the most relevant prior works below, relating them to our approach.

\textbf{Large language models.}
Language models have shown impressive capabilities in text generation~\citep{ghazvininejad-etal-2017-hafez, li2017learning, holtzman-etal-2018-learning, radford2019language, yang-klein-2021-fudge}, translation~\citep{gu-etal-2017-trainable}, question answering~\citep{pyatkin2022reinforced}, summarization~\citep{paulus2017deep, wu2018learning, bohm-etal-2019-better}, and code generation~\citep{codex,zhong2023study}.
Transforming LLMs into interactive agents capable of multi-step decision-making has been a recent focus of research. These embodied agents interact with environments by taking actions, observing responses, and adapting their strategies accordingly. 
Applications include robot learning ~\citep{singh2022progpromptgeneratingsituatedrobot,min2022film}, web navigation ~\citep{nakano2022webgpt}, and text games involving dialogue ~\citep{yao2022react,shinn2023reflexionlanguageagentsverbal,zhou2024languageagenttreesearch}.
Our work falls in the realms of training better LLM agents by proposing a novel, more scalable approach. 

\textbf{RL for LLMs.} Reinforcement learning (RL) has been proposed to improve the performance of LLMs to much success.
The most notable has been to use policy gradient methods to fine-tune LLMs from human or AI feedback ~\citep{bai2022constitutional,ouyang2022training,touvron2023llama}.
Such online methods have also been applied to train LLM agents in interactive settings, such as web search ~\citep{nakano2022webgpt} and embodied navigation ~\citep{zhai2024finetuninglargevisionlanguagemodels}. 
In contrast, offline RL methods have also been used to train LLM agents without the need for online samples ~{\citep{snell2023offline, qtransformer}.
However, RL methods require a high number of samples, and are notoriously difficult to train effectively; therefore, multi-turn RL has not been successfully scaled to frontier LLMs such as \texttt{GPT-4}~\citep{OpenAI2023GPT4}. 
We also employ multi-turn RL, but we make several key innovations that make our approach much more scalable: (1) we consider RL over high-level strategies rather than environment actions; (2) we consider goal-conditioned RL where rather than learning values, we learn likelihoods of reaching particular goal states; and (3) we do not directly fine-tune an LLM, but rather learn a light-weight auxiliary value function that can be used to guide the search of the base LLM agent using only inference APIs. These innovations make our method practical to use for frontier LLMs.

\textbf{Reasoning and planning with LLMs.}
The latest frontier models have shown that strong reasoning and planning capabilities can be elicited by different prompting techniques ~\citep{wei2023chainofthought,kojima2022largelanguagemodelszeroshot}. Such techniques are central to the design of modern LLM agents. 
Notably, ReAct leverages chain-of-thought prompting~\citep{wei2023chainofthought} to allow LLM agents to generate a composite of reasoning and task-specific actions ~\citep{yao2022react}.
However, ReAct relies on the base intuition of LLMs, which fails in more complex tasks such as web search or dialogue.
More recent approaches have proposed self-supervised methods for LLM agents to enhance their reasoning capabilities ~\citep{shinn2023reflexionlanguageagentsverbal,madaan2023selfrefineiterativerefinementselffeedback,zhou2024languageagenttreesearch}. 
Reflexion adds a step of self-reflection to ReAct allowing LLM agents to refine their initial reasoning after some environment feedback ~\citep{shinn2023reflexionlanguageagentsverbal}.
State-of-the-art methods employ tree-based search such as Monte Carlo tree search (MCTS) to guide the reasoning and planning of LLM agents \citep{yu2023promptbasedmontecarlotreesearch,zhou2024languageagenttreesearch,putta2024agentqadvancedreasoning}. 
These methods improve upon ReAct by leveraging the knowledge gained from search to improve the LLM agent's reasoning capabilities. 
However, we posit that limited search is not sufficient for LLM agents to make \emph{data-driven} decisions; existing approaches are limited by the amount of data agents can observe due to the prohibitive cost of inference-time search. 
In our work, we use offline RL to train value functions that can guide reasoning without the need for search during inference, resulting in efficient and data-driven agents.
In contrast to explicit search, another popular paradigm for improving reasoning is self-refinement~\citep{huang2023improve}, where an LLM assesses its own generations intrinsically. Our method also involves refinement, but via an auxiliary goal-conditioned value function.

\section{Preliminaries}
\vspace{-0.1in}
\label{sec:preliminaries}
\textbf{Markov decision processes.} 
We use the formalism of the Markov decision process (MDP), given by a tuple $M = (\gS, \gA, P, r, \rho, \gamma)$, where $\gS$ is the state space, $\gA$ is the action space, $P$ is the transition function, $r$ is the reward function, $\rho$ is the initial state distribution, and $\gamma$ is the discount factor. When action $a \in \gA$ is executed at state $s \in \gS$, the next state is sampled $s' \sim P(\cdot | s, a)$, and the agent receives reward $r$ with mean $r(s, a)$. Note that in many interactive settings, the state is actually partially observed, resulting in a POMDP instead of a MDP. For example, in dialogue, the state consists of the internal thoughts and intentions of other participants than the agent. However, prior analyses has shown that the two definitions can be made equivalent by instead defining states in the MDP as histories of observations so far, or in dialogue, the history of all utterances thus far \citep{Timmer2010}. 

\textbf{LLM agents in MDPs.}
Tasks considered by LLM agents can be defined under this formalism as follows. At timestep $t$, the agent \emph{state} $s_t$ of the MDP consists of the history of interaction thus far. In a dialogue, this will include the agent's past utterances, as well as responses by all other participants. Following the ReAct agent by ~\citet{yao2022react}, the agent \emph{action} $a_t$ is actually composite, consisting of a \emph{thought} $a_t^\mathsf{tht}$ where the agent performs a reasoning step, followed by the actual environment action $a_t^{\mathsf{env}}$. In dialogue, the thought could be the high-level strategy or plan the agent aims to execute, whereas the environment action is the actual utterance by the agent. 
Then, the transition function appends the environment action $a_t^{\mathsf{env}}$ as well as new observations by the environment $o_{t}$, such as responses by other interlocutors, to form the next state $s_{t+1}$. 

\textbf{Reinforcement learning.}
The objective of RL is to find a policy $\pi$ that maximizes the expected discounted return $\E{\tau \sim p^\pi(\tau)}{\sum_{t = 0}^{T-1}\gamma^t r(s_t, a_t)}$ in an MDP, where $\tau = (s_0, a_0, s_1, a_1, \hdots, s_T)$ and $p^\pi(\tau) = \rho(s_0)\prod_{t=0}^{T-1} \pi(a_t | s_t) P(s_{t+1} | s_t, a_t)$. 
The Q-function $Q^*(s, a)$ represents the discounted long-term reward attained by executing $a$ given state $s$ and then behaving optimally afterwards.
$Q^*$ satisfies the Bellman optimality equation: 
\begin{align*}
Q^*(s, a) \!=\! r(s, a) \!+\! \gamma \E{s' \sim P(\cdot|s, a)}{\max_{a'} Q(s', a')}\,.
\end{align*}
A policy can be extracted from the Q-function, i.e. $\pi^*(a | s) = \mathrm{1}[a = \arg\max_{a'}Q^*(s, a')]$.
In offline RL, we want to estimate Q-function from a dataset $\mathcal{D}$ comprising of samples $(s, a, s', r)$, where each sample corresponds to one timestep of interaction, and $a$ come from a behavior policy $\pi_\beta(\cdot|s)$ (which might correspond to a mixture of multiple policies). 
\citet{kostrikov2021offline} propose Implicit Q-learning (IQL), which trains Q-values $\hat{Q}(s, a)$ and state-values $\hat{V}(s)$ with the following loss functions: 
{\small
\begin{align*}
L_Q \!=\! \mathbb{E}_{(s, a, s', r) \sim \mathcal{D}}\!\left[\left(r + \gamma \hat{V}(s') - Q(s, a)\right)^2\right]\,, \quad
L_V \!=\! \mathbb{E}_{(s, a) \sim \mathcal{D}}\!\left[L_2^\tau\left(\hat{Q}(s, a) - V(s, g)\right)\right]\,,
\end{align*}
}
where $L_2^\tau$ is the expectile loss with hyperparameter $\tau \in [0.5, 1)$.

\section{\textsf{PNLC}: Planning with a Natural Language Critic}
\label{sec:method}

Here, we describe our proposed method, which we dub Planning with a Natural Language Critic (\textsf{PNLC}).
PNLC provides a general approach for training LLM agents, using only a small prior dataset of trajectories.
During the training phase, our method uses this dataset to train a goal-conditioned value function that can predict the probability of different interaction outcomes given the agent's proposed action. At inference time, we utilize this goal-conditioned value function as a natural language critic, which helps an LLM agent refine its strategy thought at each step of interaction, via querying the value function for the probability of different outcomes and using these probabilities to perform self-refinement over possible thoughts.

An overview of our approach can be found in Figure~\ref{fig:method}.
During offline training, we aim to learn a goal-conditioned value function $Q(s, a^{\mathsf{tht}}, g)$. Here, the goal $g$ is a possible future state that can result from following thought $a^{\mathsf{tht}}$ in state $s$, and $Q(s, a^{\mathsf{tht}}, g)$ denotes the likelihood of reaching the goal future state $g$. 
Then, during inference, we can evaluate thoughts by an LLM agent by analyzing possible futures. Specifically, by sampling possible positive and negative future states and providing their likelihoods, an LLM agent can self-refine its original thought. We describe each stage in greater detail below, but defer specifics such as training hyperparameters and prompts used to Appendix~\ref{app:imp_details}.

\subsection{Training a Goal-Conditioned Value Function}
\begin{figure*}
\centering
\includegraphics[width=\linewidth]{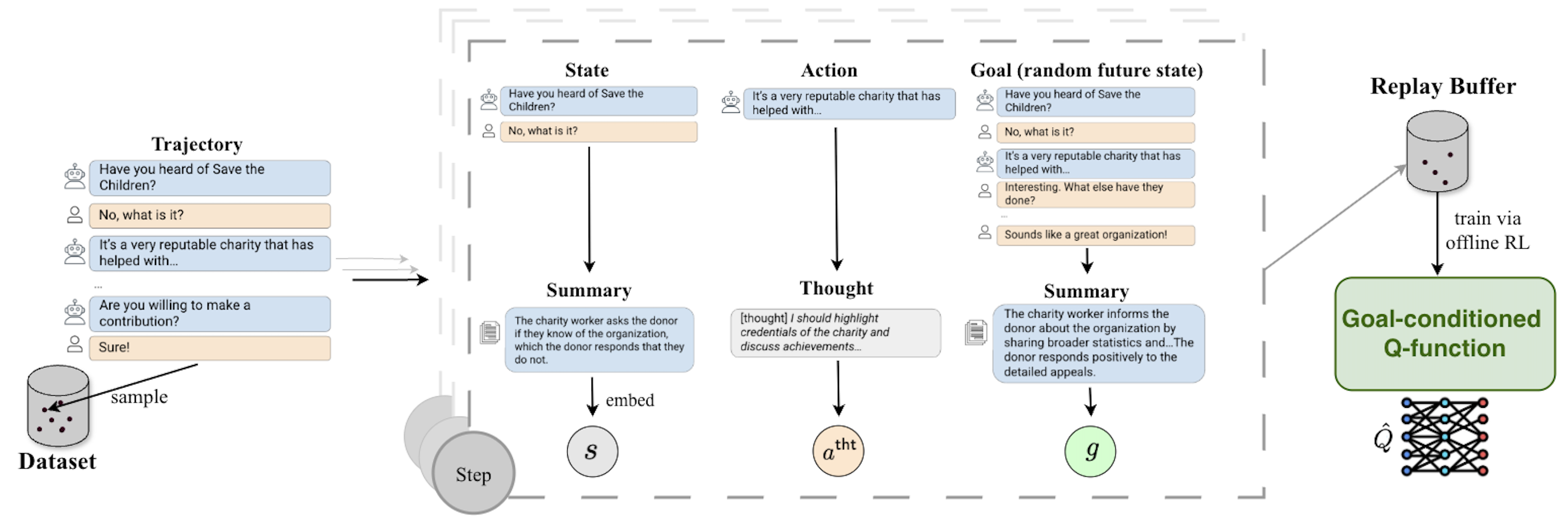}
\caption{During \emph{offline training}, training samples are summarized and embedded, and a goal-conditioned value function (which is just a small MLP) is trained over the embeddings.}
\label{fig:method_training}
\end{figure*}
Our proposed natural language critic relies on a goal-conditioned value function to evaluate actions in lieu of inference-time search. As input, we require a dataset of task-specific trajectories $\mathcal{D}$ by any prior agent. The agent does not need to be near-optimal, but only exhibit a variety of different, even incorrect, strategies ~\citep{kumar2022prefer}. In our experiments, we use an LLM agent with chain-of-thought prompting, which are known to exhibit suboptimal performance in tasks requiring long-term planning. 

Recall that we evaluate only the thought component of the agent action. 
This is because we view the thought as an abstraction of the full action that contains all the important information to plan over. For example, in Figure~\ref{fig:dialogue_example}, the high-level thought, which was to appeal to the credibility of the organization, captures the semantics of the agent's next utterance without the noise of linguistic structure or syntax.

From each trajectory in the dataset we extract training samples $(s, a^{\mathsf{tht}}, s', g)$ consisting of state, thought, next state, and a goal that is a random future state in the trajectory. 
Note that following the formalism in Section~\ref{sec:preliminaries}, the states are all histories of interaction with the environment, and a goal adds possible future interactions to the original state.
To make learning more efficient, we consider the following simplifications to the training process:

\begin{enumerate}
\item[(1)]\emph{Summarizing trajectories:} Rather than considering full histories of interaction, we instead summarize them into compact descriptions. Though the contents of the summaries are task-specific, we design them to capture all the relevant information needed to make decisions regarding the next thought or strategy to adopt. For example, in dialogue, the summarizations aim to describe the thoughts of the participants in the discussion history rather than their actual utterances. We provide example summarization prompts in Appendix~\ref{app:imp_details}.

\item[(2)]\emph{Embedding language descriptions:} Finally, rather than learning value functions over natural language, either as thoughts or summaries of histories, we instead consider their low-dimensional embeddings that are output by an LLM. Because LLMs have a basic grasp of how to parse natural language, we view the embeddings as sufficient to learn over. This has the added benefit of our learned value functions being lightweight architectures, rather than full Transformers, which are proven to be slow to train.
\end{enumerate}

The training samples are used to learn a goal-conditioned Q-value $\hat{Q}(s, a^{\mathsf{tht}}, g)$ by modifying the loss function of the IQL algorithm~\citep{kostrikov2021offline} to condition on goals:
{\small
\begin{align*}
L_Q \!=\! \mathbb{E}_{(s, a^{\mathsf{tht}}, s', g) \sim \mathcal{D}}\!\left[\left(r(s, g) \!+\! \gamma \hat{V}(s', g) \!-\! Q(s, a^{\mathsf{tht}}, g)\right)^2\right]\,,\,
L_V \!=\! \mathbb{E}_{(s, a^{\mathsf{tht}}, g) \sim \mathcal{D}}\!\left[L_2^\tau\left(\hat{Q}(s, a^{\mathsf{tht}}, g) - V(s, g)\right)\right]\,,
\end{align*}
}
where $r(s, g)$ is a binary indicator for whether state $s$ is the goal state $g$. At convergence, $\hat{Q}(s, a, g)$ should predict the likelihood of reaching goal $g$ after taking action $a$ at state $s$.

An illustration of the offline stage of our approach can be found in Figure~\ref{fig:method_training}. Note that though the summarization and embeddings require calls to an LLM, because they do not require particularly sophisticated reasoning, we are able to use cheaper LLM models to perform this step. In practice, we use GPT-3~\citep{OpenAI2022ChatGPT} to extract embeddings, rather than GPT-4~\citep{OpenAI2023GPT4} that is used in the final decision-making process. Also note that because our goal-conditioned Q-function does not take natural language inputs but rather embeddings, our learned Q-function is simply a MLP with 2 hidden layers rather than a full Transformer. 

\subsection{Planning using a Natural Language Critic}

\begin{figure*}
\centering
\includegraphics[width=\linewidth]{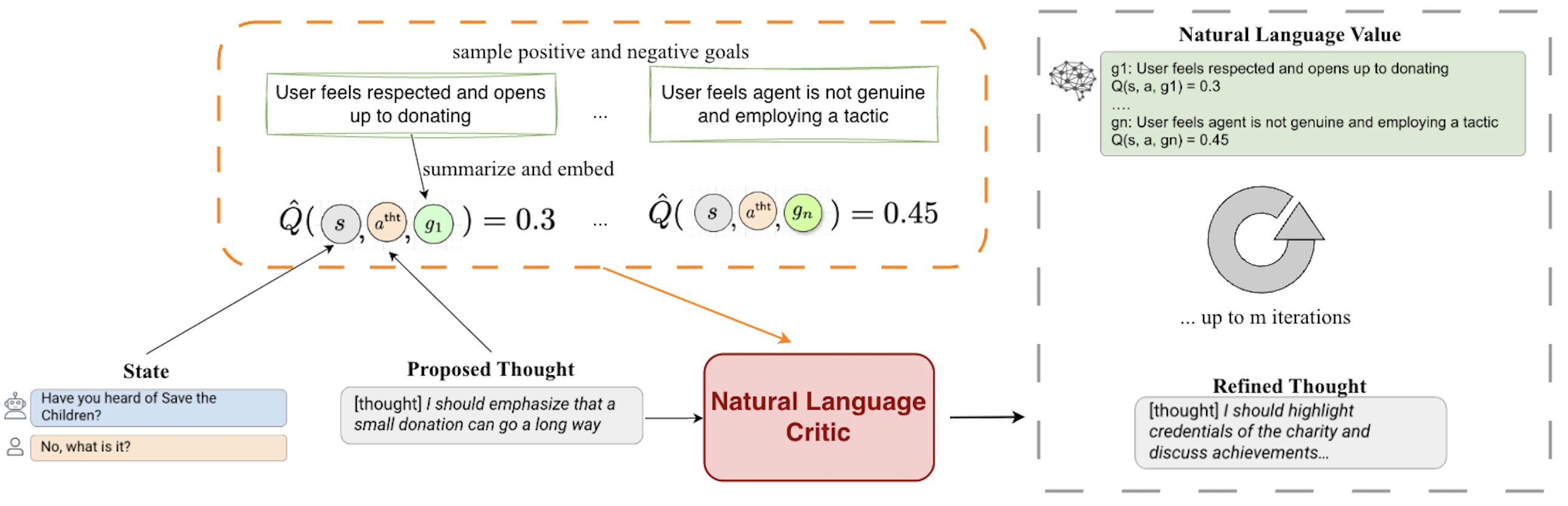}
\caption{During \emph{inference}, a natural language critic uses the value function to produce an informative analysis of possible future outcomes. This natural language value is used by the LLM agent to refine its proposed reasoning.}
\label{fig:method_inference}
\vspace{-0.2in}
\end{figure*}
Now, we describe how our LLM agent leverages the goal-conditioned value function learned offline as a natural language critic for complex decision-making.
Naively, the LLM agent could simply use the value function to take the action with the highest probability of reaching a desired goal.
However, LLM agents historically perform better when allowed to refine their actions using feedback ~\citep{shinn2023reflexionlanguageagentsverbal,madaan2023selfrefineiterativerefinementselffeedback}.
This is typically done post-hoc using trajectories from earlier attempts or from simulation.
We instead use the natural language critic as a way to provide feedback over proposed actions,
so that the LLM agent can refine its actions without the need to gather any trajectories during inference. 

Specifically, the natural language critic takes state $s$ and proposed thought $a^{\mathsf{tht}}$, and performs the following operations. First, a LLM generates $n$ hypothetical positive and negative goals $(g_1, g_2, \hdots, g_n)$ based on the current state. Each goal represents a possible future outcome of the agent's proposed thought. Then, we feed the quantities, including the sampled goals, to our learned goal-conditioned value function, which for each goal outputs a scalar quantity that is the likelihood of reaching the goal. The resulting goals $(g_1, \hdots, g_n)$ and corresponding values $(\hat{Q}(s, a^{\mathsf{tht}}, g_1), \hdots, \hat{Q}(s, a^{\mathsf{tht}}, g_n))$ are combined into a natural language value. 
The natural language value, compared to scalar-based values, is much richer in information and also greatly improves interpretability. 
Furthermore, the inclusion of negative outcomes allows the LLM to better correct its proposed thoughts. 

The feedback provided by the natural language critic can be viewed as a compact summarization of a plethora of possible future trajectories, much more than can be obtained via inference-time search.
Hence, the natural language critic can co-opt the traditional search procedure in traditional LLM reasoning processes. 
Namely, given a proposed thought and natural language value, the LLM agent is optionally allowed to suggest an improvement to its original thought, up to a total of $m$ iterations of refinement. 
In practice, we find that $m\!=\!2$ with only one additional round of refinement was sufficient to achieve good performance across our evaluated tasks. 
See Figure~\ref{fig:method_inference} for an illustration of the inference stage of our approach. 
We will show later that our method not only profits from efficiency but is also more effective than existing approaches across a variety of benchmarks.  

\section{Experiments}

To demonstrate the effectiveness of PNLC, we evaluate our method on a variety of multi-turn LLM agent benchmark tasks: web shopping ~\citep{yao2023webshop}, social deduction games ~\cite{light2023avalonbenchevaluatingllmsplaying}, and persuasion ~\citep{wang2019persuasion}. 

\subsection{Task Descriptions}
\label{app:tasks}
All tasks we evaluate on are illustrated in Figure~\ref{fig:tasks}. In this section, we provide descriptions for each task, as well as how training data was generated.

\textbf{Web Shopping.} WebShop~\citep{yao2022react} is an online shopping environment where an agent processes unstructured text data (in the form of descriptions crawled from Amazon) to purchase a product given some initial user specifications. At the end, the agent receives a reward between $0$ and $1$ depending on the similarity between the purchased and ground-truth desired item.
The performance is gauged using two metrics:
the score, reflecting the average reward received, and success rate,
indicating the frequency with which the chosen product fulfills all specified conditions. 
All relevant methods train on the same dataset consisting of $12$k initial user instructions, of which we randomly held out $100$ for evaluation. With the remaining instructions, we generate a dataset of trajectories in which we simulate a suboptimal agent by prompting \texttt{GPT-3.5} with few-shot examples similar to that done by ~\citet{yao2022react}.

\textbf{Social Deduction.} AvalonBench~\citep{light2023avalonbenchevaluatingllmsplaying} is a testbed for LLM agents modeled after the team-based discussion game Resistance Avalon, where agents must jointly deceive other agents while deducing their roles, all through natural language dialogue. Each game consists of 5 players, 3 of which are on the good side and the remaining 2 evil. A notable asymmetry of this game is that all but one of the good players do not know other players’ identities, while evil players are aware of who their teammates are. The one exception is the role \emph{Merlin}, which the LLM agent will play due to being the hardest role in the game. Merlin knows the identity of the evil players, and must guide the other good players whilst maintaining his own identity a secret; if his identity is discovered by the evil players, they are allowed to assassinate him, and the evil side wins the game. 
The LLM agent wins if the good side succeeds in at least 3 of 5 rounds, where a successful round means a subset of players chosen by a designated leader for the round all vote to accomplish a mission. \citet{light2023avalonbenchevaluatingllmsplaying} implemented baseline agents for all roles in the game, which use \texttt{GPT-3.5}~\citep{OpenAI2022ChatGPT} for discussion and a heuristic for action selection. We simulate 2.5k trajectories using these baseline agents, in which the baseline Merlin achieves a $21.4\%$ win rate.

\textbf{Persuasion.} We consider a goal-oriented dialogue task, inspired by~\citet{wang2019persuasion}.
In this task, an LLM agent must persuade a participant to donate to Save the Children, a non-governmental organization dedicated to international assistance for children.
At the end, the participant can choose to donate up to \$2 total. Because the donation amount is a very noisy signal, following prior work ~\citep{phuong2024evaluatingfrontiermodelsdangerous}, we model potential donors by prompting \texttt{GPT-3.5}~\citep{OpenAI2022ChatGPT} to generate responses. 
To ensure that intelligent persuasion is required to complete the task successfully, the simulated donor is prompted to be skeptical of donating to charities. In our setup, the simulated user interacts with the LLM agent for up to 10 turns of dialogue, then must choose an amount to donate. We college 2.5k dialogues using naive prompting of \texttt{GPT-3.5}~\citep{OpenAI2022ChatGPT} as the data collection agent. The average donation amount in the dataset is just $\$0.21$.

\begin{figure}
\centering
\includegraphics[width=\linewidth]{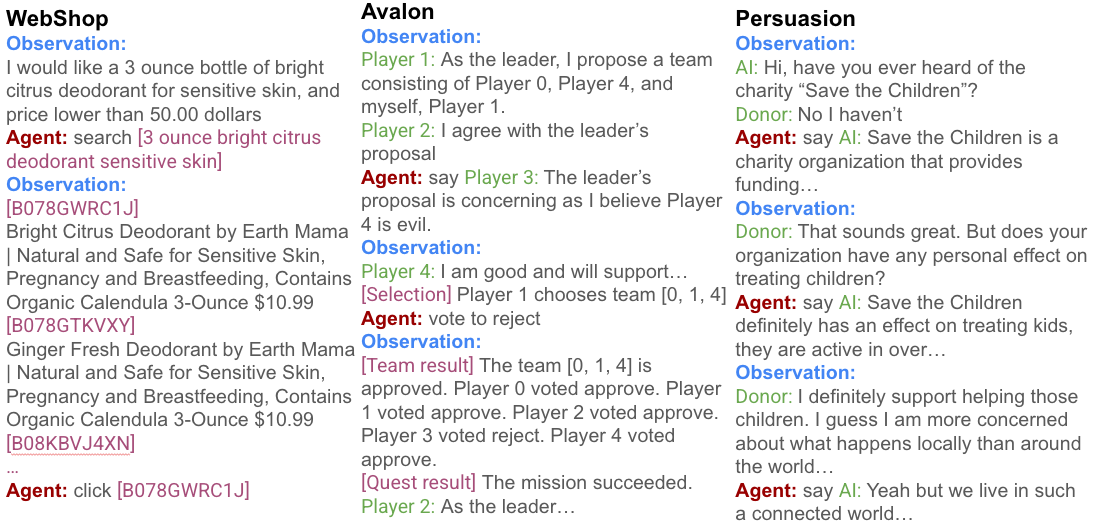}
\caption{The various tasks we consider, spanning tool use, games, and goal-oriented dialogue.}
\label{fig:tasks}
\vspace{-0.1in}
\end{figure}

\subsection{Evaluated Methods}

For each task, we run PNLC with $n\!=\!4$ goals and a single round $m\!=\!2$ of refinement. We chose $n\!=\!4$ to have a mixture containing $2$ positive and negative goals. We evaluate against a range of fine-tuning and prompting baselines; some of these prior methods can be evaluated on the full suite of benchmark tasks, while others are designed specifically for one of the domains. The algorithms can be grouped into the following $3$ classes:

\textbf{RL fine-tuning.} To compare to a prior method that uses multi-turn RL, we use two variants of the recently proposed ArCHeR algorithm~\citep{zhou2024archertraininglanguagemodel}, which uses an actor-critic architecture to directly fine-tune LLMs for interactive tasks. We compare to two variants of a popular fine-tuning algorithm: (1) offline ArCHer and (2) online ArCHer, which employs interactive rollouts through the environment to further optimize the agent's behavior. Due to the complexity of end-to-end value-based RL methods, we run ArCHeR using the smaller \texttt{GPT-2} model~\citep{radford2019language} to represent the policy, following prior work~\citep{zhou2024archertraininglanguagemodel}.

\textbf{LLM reasoning.} Next, we evaluate a variety of sophisticated strategies used to invoke reasoning and planning capabilities on a \texttt{GPT-4} LLM agent~\citep{OpenAI2023GPT4}, the most complex of which performs inference-time search. We compare to: (1) ReAct~\citep{yao2022react}, which uses chain-of-thought prompting for LLM agents to reason about their actions solely using the zero-shot capabilities of the LLM; (2) Reflexion ~\citep{shinn2023reflexionlanguageagentsverbal}, which adds $n\!=\!5$ additional simulations of full trajectories that the LLM agent is allowed to refine with; (3) LATS ~\citep{zhou2024languageagenttreesearch}, which performs full MCTS simulation and selects its actions based on the results of the search. We evaluate both a fast and slow version of each method, with either $n\!=\!5$ or $n\!=\!30$ simulations, to provide a fairer point of comparison for our method (the fast version requires a similar inference budget as ours). 
Finally, we evaluate a custom adaptation of \citet{huang2023improve} from question-answering to multi-turn tasks called (4) Self-Plan; rather than using majority vote on $n$ answers by the model, we instead have the model select from $n\!=\!5$ of its own generations. In contrast to LATS, Self-Plan does not perform any search, but performs self-refinement by assessing its own generations.

\textbf{Task-specific design.} A number of prior works propose task-specific methods that may involve a mixture of RL fine-tuning and inference-time search. Such methods typically leverage domain-specific knowledge, such as a taxonomy of high-level but task-specific strategies, which do not directly generalize to other tasks. We select and compare against one such method for each task that achieves state-of-the-art performance: 

\begin{enumerate}
\item[(1)] \emph{WebShop:} \citet{putta2024agentqadvancedreasoning} achieved state-of-the-art performance with their method Agent Q, which mixes offline fine-tuning with inference-time search. In the offline phase, rather than value-based RL, Agent Q uses DPO ~\citep{rafailov2023direct} on preference pairs of trajectories in the dataset. Then, during inference, Agent Q performs MCTS search to generate $n$ simulations before selecting an environment action. Like with LATS, we consider fast $n\!=\!5$ and slow $n\!=\!30$ variations of Agent Q. Note that because Agent Q requires modifying the weights of the base LLM agent, they train on \texttt{Mixtral-7B}~\citep{jiang2024mixtralexperts} as more powerful models such as \texttt{GPT-4}~\citep{OpenAI2023GPT4} only expose inference APIs.

\item[(2)] \emph{AvalonBench:} Strategist~\citep{light2024strategistlearningstrategicskills} proposes a hierarchical prompting mechanism that additionally uses MCTS via population-based self-play to propose and refine high-level strategies. The high-level strategies searched over by Strategist are similar to the thoughts refined by our proposed method, but ours does not rely on any self-play during inference. Again, to keep a fair comparison with inference time, we consider both a fast $n\!=\!5$ and slow $n\!=\!30$ variations of Strategist varying the number of simulations during search. 

\item[(3)] \emph{Persuasion:} \citet{yu2023promptbasedmontecarlotreesearch} propose GDP-Zero, which prompts the LLM to perform tree-search over possible high-level strategies at every timestep, simulating responses by both interlocutors in the dialogue, then selects the best action according to the search.
This method is similar to LATS, but searches over a task-specific persuasion taxonomy rather than the full space of actions. We consider $n\!=\!5$ and $n\!=\!30$ varations of the approach, varying the number of simulated dialogues.
\end{enumerate}

\begin{table}
\centering
\begin{tabularx}{0.75\textwidth}{l c c c c}
\toprule
 & \multicolumn{2}{c}{WebShop}
 & \multicolumn{1}{c}{Avalon}
 & \multicolumn{1}{c}{Persuasion}
 \tabularnewline \cmidrule(l){2-3} 
Method & Score & SR & Winrate & Avg. Donation \\
\hline\hline 
      ArCHer ~\citep{zhou2024archertraininglanguagemodel} & $62.3$ & $32.0$ & $19.0$ & $0.36$\\
      Offline ArCHer ~\citep{zhou2024archertraininglanguagemodel} & $57.3$ & $28.0$ & $18.0$ & $0.31$ \\
      \hline
      ReAct ~\citep{yao2022react} & $55.1$ & $27.0$ & $21.0$ & $0.54$ \\
      Reflexion ~\citep{shinn2023reflexionlanguageagentsverbal} & $60.8$ & $29.0$ & $26.0$ & $0.54$\\
      LATS ($n\!=\!30$) ~\citep{zhou2024languageagenttreesearch} & $74.9$ & $44.0$ & $38.0$ & $0.78$\\
      LATS ($n\!=\!5$) ~\citep{zhou2024languageagenttreesearch} & $53.9$ & $28.0$ & $22.0$ & $0.52$\\
      Self-Plan ~\citep{huang2023improve} & $54.8$ & $28.0$ & $27.0$ & $0.55$\\
      \hline
      Agent Q ($n\!=\!30$) ~\citep{putta2024agentqadvancedreasoning} & $77.1$ & $\mathbf{48.0}$ & -  & - \\
      Agent Q ($n\!=\!5$) ~\citep{putta2024agentqadvancedreasoning} & $63.2$ & $35.0$ & -  & - \\
      Strategist ($n\!=\!30$) ~\citep{light2024strategistlearningstrategicskills} & -  & - & $42.0$ & - \\
      Strategist ($n\!=\!5$)~\citep{light2024strategistlearningstrategicskills} & -  & -  & $31.0$ & - \\
      GDP-Zero ($n\!=\!30$) ~\citep{yu2023promptbasedmontecarlotreesearch} & -  & - & - & $0.74$\\
      GDP-Zero ($n\!=\!5$) ~\citep{yu2023promptbasedmontecarlotreesearch} & -  & - & - & $0.47$ \\      
      \hline
      PNLC (ours)& $\mathbf{78.2}$ & $\mathbf{48.0}$ & $\mathbf{47.0}$ & $\mathbf{0.87}$ \\
      \bottomrule
\bottomrule
\end{tabularx}
\caption{Mean evaluation results across all methods and baselines over 100 independent test runs. Across the board, our method PNLC performs best while only using less than 10$\%$ the inference budget of the next best approaches.}
\label{tab:results}
\vspace{-0.2in}
\end{table}

\subsection{Results} 
A table of all results can be found for all considered benchmarks in Table~\ref{tab:results}.
We see that PNLC performs best across all tasks, the closest competitors being task-specific approaches such as Agent Q in WebShop or Strategist in Avalon. However, it is important to note that such methods use domain-specific prompting mechanisms
and do not naively generalize to other tasks; in addition, Agent Q requires DPO fine-tuning on a \texttt{Mixtral-7B} LLM, which is much more expensive than the lightweight value function that we train. Furthermore, the $n\!=\!30$ version of search approaches requires around $10\times$ more time to make a single decision ---
$46$s for Agent Q compared to $5$s for ours in WebShop, and $62$s for Strategist compared to $6$s for ours in Avalon. 
Interestingly, though RL fine-tuning approaches such as ArCHer train on the same data as ours, but actually often perform the worst. We attribute this to the fact that standard RL fine-tuning methods rely on a much weaker LLM as the agent; qualitatively, such LLM agent often gives nonsensical or off-topic responses.

We give qualitative examples of how our method refines decisions in Figure~\ref{fig:planning_examples}. In Avalon, we see that our method corrects the base LLM agent who was initially going to accuse a player of being evil, which may inadvertently reveal their role to the evil players. Furthermore, in persuasion, the LLM agent using our method changes their strategy to directly address the skepticism in the potential donor, rather than continuing to ineffectively tout the charity's past achievements. 

Note that our experimental evaluation considers a diverse variety of methods that use different underlying LLMs as policies, ranging from large modern models such as \texttt{GPT-4}, to medium open-source models such as \texttt{Mixtral-7B}, to small older models such as \texttt{GPT-2}. Unfortunately, rigorously controlling for model size is impractical in such experiments, and wherever possible we chose models that match those actually used in the prior works that we compare to. However, it is worth noting that despite using the largest possible frontier model in \texttt{GPT-4}, our method still ends up being faster at inference time than many of the methods we compare to. Nonetheless, this heterogeneity represents a limitation of our experiments.

\subsection{Ablation Study}
We conducted an ablation analysis to better understand where the improvements in our method were coming from. We test two important components of our method: (1) offline goal-conditioned training to obtain natural language values, and (2) a novel inference procedure of refining thoughts using these values. 

To evaluate the efficacy of each component, we evaluate two modifications of PNLC. The first removes the goal-conditioning, instead learning a scalar Q-value conditioned only state and thought, which simply predicts likelihood of success in the task. Recall that we use goals in our natural language critic in order to create informative descriptions that are more naturally used during LLM reasoning. Now, the LLM agent sees a scalar value instead of natural language, and makes its refinement off that value.
The second modification replaces the inference-time refinement procedure with more traditional best-of-$n$ selection~\citep{li2024QProbe,sessa2024bondaligningllmsbestofn}. Instead of being given the chance to refine its original thought using feedback provided by the natural language critic, this method instead has the LLM agent generate $n\!=\!5$ thoughts, from which one is chosen by the same agent. Note that our refinement procedure was designed with the hypothesis that an LLM agent can produce better thoughts \emph{after} observing what may go wrong, which this modification cannot leverage.  

Finally, we include an ablation dubbed ReAct+Replan that does not affect the inference pipeline at all, so the same self-refinement inference method using subgoals is preserved, but the scores of subgoals are directly inferred by the frontier LLM rather than learned via offline goal-conditioned RL. This ablation tests whether any training is necessary, or whether the scaffolding we propose can be handled entirely using the base capabilities of frontier LLMs.

We show results in Table~\ref{tab:ablation}. We notice that the first two ablations perform comparably to other LLM reasoning baselines such as ReAct that rely only on prompting,
meaning the LLM agent does not effectively use the auxiliary learned value function. This shows that both components of our approach --- the ability to (1) craft natural language descriptions via goals and (3) refine thoughts using such descriptions --- are both crucial to the success of our algorithm. 
Finally, ReAct+Replan performs better but still noticeably worse than our method, which shows that offline training is important, and that even frontier LLMs are incapable of providing accurate scores for goals in long-horizon tasks a priori. Qualitatively, we see that even frontier LLMs provide uncalibrated, often overly-optimistic scores, such that they rarely deem their own actions to be suboptimal. In this manner, the refinement stage in ReAct+Replan often only decides that the initial action was good.

\begin{figure}
\centering
\includegraphics[width=\linewidth]{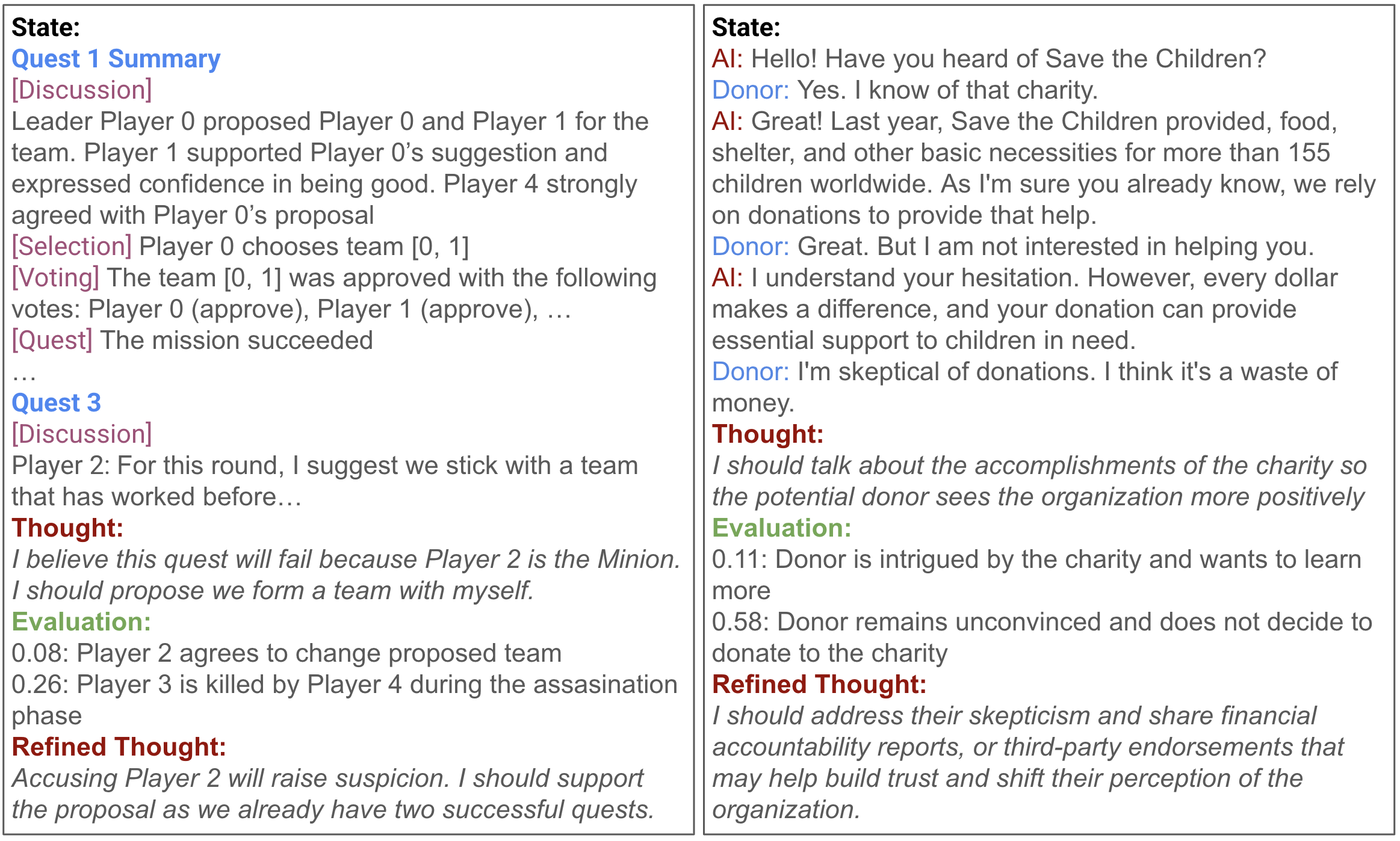}
\caption{Example planning steps by our method. \emph{Left:} In the social deduction task (Avalon), the agent originally intended to reveal the role of Player 2, which would have raised suspicion. After refinement, the agent keeps the information hidden. \emph{Right:} In the persuasion task, the agent learns to address skepticism, better sharing how the charity takes accountability rather than focusing on their existing accomplishments.}
\label{fig:planning_examples}
\end{figure} 

\begin{table}
\centering
\begin{tabularx}{0.75\textwidth}{l c c c c}
\toprule
 & \multicolumn{2}{c}{WebShop}
 & \multicolumn{1}{c}{Avalon}
 & \multicolumn{1}{c}{Persuasion}
 \tabularnewline \cmidrule(l){2-3} 
Method & Score & SR & Winrate & Avg. Donation \\
\hline\hline 
     ReAct ~\citep{yao2022react} & $55.1$ & $27.0$ & $21.0$ & $0.54$ \\
     \hline
      PNLC (w/o goals) & $55.4$ & $27.0$ & $25.0$ & $0.53$ \\
      PNLC (w/o refinement) & $55.6$ & $30.0$ & $28.0$ & $0.61$  \\
      \hline
      ReAct+Replan & $59.1$ & $32.0$ & $22.0$ & $0.62$ \\
      \hline
      PNLC &  $\mathbf{78.2}$ & $\mathbf{48.0}$ & $\mathbf{47.0}$ & $\mathbf{0.87}$ \\
\bottomrule
\end{tabularx}
\caption{Mean evaluation results for ablations across 100 independent test runs. We see that all ablations of our method perform much worse, meaning that both the goal-conditioning and self-refinement process are important to our method. Without either, the ablations match the performance of simple chain-of-thought prompting.}
\label{tab:ablation}
\vspace{-0.25in}
\end{table}


\section{Discussion}

In this paper, we propose a new method \textsf{PNLC} that imbues LLM agents with long-term planning and reasoning capabilities, using multi-turn RL in an \emph{efficient} and \emph{scalable} manner. To our knowledge, our approach is the first RL method that scales to frontier LLMs as agents. The key insight that we leverage is to not directly train a LLM policy, but an auxiliary, light-weight goal-conditioned value function.
This value function can be used at inference time to assess and refine the thoughts of any LLM agent. As shown in our experiments, our approach also significantly outperforms state-of-the-art search techniques in both compute time and overall performance.

\textbf{Limitations.}
The most notable limitation of our approach is the need to train task-specific value functions. Though our value functions are light-weight and easy to train, an important direction of future research is the adaptation of our method to allow generalist reasoning in LLM agents. Furthermore, another limitation of our method is the reliance on an LLM's intuition to reason about possible future states. This may be hard to do for tasks that are out-of-domain of an LLM's base reasoning capabilities. For example, it may be risky to rely on an LLM's ``intuition'' for tasks like medical diagnosis, which require a plethora of domain-specific knowledge. Finally, another important avenue of future research is an investigation of the quality and diversity of the offline data necessary for our approach to work well. 


\bibliography{paper}
\bibliographystyle{plainnat}

\newpage 

\appendix

\section{Implementation Details}
\label{app:imp_details}

In this section, we provide details of implementation of our method across the various benchmarks we consider. Details include the prompts used during inference, as well as hyperparameters configured during offline training. For any task, implementing our method requires crafting prompts for $3$ different components that involve natural language: (1) summarizing the states, (2) generating positive and negative goals, and (3) refining thoughts. 

\subsection{WebShop Prompts}

{\centering
\fbox{
\begin{minipage}{\linewidth}
\textbf{Summarization prompt:} \\
Below is an instruction for an item and the prefix of a trajectory that aims to buy an item that exactly matches the specification: \\
\textcolor{blue}{[INSTRUCTION+TRAJECTORY]} \\
Please summarize the trajectory prefix. Try to keep all the useful information, including actions taken and important observations. \\

Here are examples below: \\
\textcolor{blue}{[INSTRUCTION+TRAJECTORY+SUMMARY]}
\end{minipage}
} \par}

\paragraph{Example summary:} The agent did a search for “variety pack of chips” but most results did not meet the dairy free nor the \$30 budget. The agent clicked on a variety pack item that is \$100. \medskip

{\centering
\fbox{
\begin{minipage}{\linewidth}
\textbf{Goal proposal prompt:} \\
Below is an instruction for an item and the prefix of a trajectory that aims to buy an item that exactly matches the specification and the current thought by the agent which hints at what action they want to take next: \\
\textcolor{blue}{[INSTRUCTION+TRAJECTORY+THOUGHT]} \\
Please propose a future \{positive|negative\} goal. The goal can either describe a page of items or a particular item that the agent may click. \\

Here are examples below: \\
\textcolor{blue}{[INSTRUCTION+TRAJECTORY+GOAL]}
\end{minipage}
} \par} 

\paragraph{Example negative goal:} The agent sees a list of products whose prices are too high. \medskip

{\centering
\fbox{
\begin{minipage}{\linewidth}
\textbf{Refinement prompt:} \\
Below is an instruction for an item and the prefix of a trajectory that aims to buy an item that exactly matches the specification: \\
\textcolor{blue}{[INSTRUCTION+TRAJECTORY]} \\
Here is the current thought by the agent and assessment of the possible futures after following this thought by a critic trained on lots of data: \\
\textcolor{blue}{[THOUGHT+NATURAL LANGUAGE VALUE]}
First, determine whether or not the current proposed thought will lead to success.
If not, in a few sentences, diagnose a possible reason for failure and devise a
new thought that aims to mitigate the same failure. 
\end{minipage}
} \par} 

\paragraph{Example original thought:} The third item is dairy-free and seems to be what I want.

\paragraph{Example refinement:} To get broader results, I will search for “variety pack of chips” and check if the new results better satisfy the constraints.

\subsection{WebShop Training Details}

\begin{table}[H]
\centering
\begin{tabular}{lrr}
\toprule
Hyperparameter & \multicolumn{2}{r}{Setting } \\
\midrule
\midrule
Hidden-layer size && 64*64 \\
IQL $\tau$ && 0.8 \\
Discount factor && 0.99 \\
Batch size && 32 \\
Target network update $\alpha$ && 0.005  \\
Number of updates per iteration && 50 \\
Number of iterations && 100 \\
Optimizer && AdamW \\
Learning rate && 4e-4 \\

\bottomrule
\end{tabular}
\end{table}

\subsection{Avalon Prompts}
For game rules, we use the prompt specified in the Appendix of \citet{light2023avalonbenchevaluatingllmsplaying}. Similarly, for generating actions for proposing teams (when chosen as leader), voting, or for discussion, we use the prompts used by Strategist as specified in the Appendix of \citet{light2024strategistlearningstrategicskills}.

{\centering
\fbox{
\begin{minipage}{\linewidth}
\textbf{Summarization prompt:} \\
\textcolor{blue}{[RULES]} \\
You are the Merlin role. Players \{evil players\} are evil and \{good players\} are good.
Below is summary and log of what happened in the previous round: \\
\textcolor{blue}{[SUMMARY+LOG]} \\
Please summarize the history. Try to keep all the useful information, including your
identification and your observations of the game. \\

Here are examples below: \\
\textcolor{blue}{[SUMMARY+LOG+NEW SUMMARY]}
\end{minipage}
} \par}

\paragraph{Example summary:} 
So far, we have completed two quests, both of which were successful. This means that
Good is currently leading with two successful quests, while Evil has not yet been able to sabotage any quests.
Player 4 seems to suspect that either me (Player 3) or Player 0 is Merlin.
In the previous round, Player 2 proposed a team consisting of themselves and Player 0,
which was successful. 
\medskip

{\centering
\fbox{
\begin{minipage}{\linewidth}
\textbf{Goal proposal prompt:} \\
\textcolor{blue}{[RULES]} \\
You are the Merlin role. Players \{evil players\} are evil and \{good players\} are good.
Below is summary of what has happened so far, as well as thoughts detailing your intended action: \\
\textcolor{blue}{[SUMMARY+THOUGHT]} \\
Please propose a future \{positive|negative\} goal. The goal should be a summary of what happens at the end of this round or one in the near future. 

Here are examples below: \\
\textcolor{blue}{[SUMMARY+THOUGHT+GOAL]}
\end{minipage}
} \par} 

\paragraph{Example positive goal:} I propose that we stick with the same composition of Player 0 and 2, which have proved successful on quests so far. By gaining the trust of Player 0 and 2, we are able to vote together and successfully complete a third mission, winning the game. \medskip

{\centering
\fbox{
\begin{minipage}{\linewidth}
\textbf{Refinement prompt:} \\
\textcolor{blue}{[RULES]} \\
You are the Merlin role. Players \{evil players\} are evil and \{good players\} are good.
Below is summary of what has happened so far: \\
\textcolor{blue}{[SUMMARY+THOUGHT]} \\
Here is your current thought on how to proceed, as well as an assessment of possible positive and negative outcomes of your decision made by a critic trained on lots of data: \\
\textcolor{blue}{[THOUGHT+NATURAL LANGUAGE VALUE]} \\
First, determine whether or not the current pro-
posed strategy will lead to success. If not, in a few
sentences, diagnose a possible reason for failure
and devise a new strategy that aims to mitigate the
same failure.
\end{minipage}
} \par} 
\paragraph{Example original thought:} As the leader this round, I believe sticking with the same team composition of Player 0 and 2 will prove successful, as I am reasonably confident they are both Good. 

\paragraph{Example refinement:} From previous discussion, Player 4 appears to suspect that I am Merlin. In case he is the Assassin, I should act more confused and defer to Player 0 to lead discussion and follow his decision.                                         

\subsection{Avalon Training Details}

\begin{table}[H]
\centering
\begin{tabular}{lrr}
\toprule
Hyperparameter & \multicolumn{2}{r}{Setting } \\
\midrule
\midrule
Hidden-layer size && 128*128 \\
IQL $\tau$ && 0.8 \\
Discount factor && 0.99 \\
Batch size && 32 \\
Target network update $\alpha$ && 0.01  \\
Number of updates per iteration && 50 \\
Number of iterations && 100 \\
Optimizer && AdamW \\
Learning rate && 8e-4 \\

\bottomrule
\end{tabular}
\end{table}

\subsection{Persuasion Prompts}

{\centering
\fbox{
\begin{minipage}{\linewidth}
\textbf{Summarization prompt:} \\
The following is some background information about Save the Children.
Save the Children is head-quartered in London, and they work to help fight poverty around the world.
Children need help in developing countries and war zones. Small donations like \$1 or \$2 go a long way to help. \\

The following is a conversation between a Persuader and a Persuadee about a charity called Save the Children. The Persuader is trying to persuade the Persuadee to donate to Save the Children. \\
\textcolor{blue}{[DIALOGUE]}\\
Please summarize the dialogue. Try to keep all the useful information, including strategies employed by the Persuader and how the Persuadee responded. \\

Here are examples below: \\
\textcolor{blue}{[DIALOGUE+SUMMARY]}
\end{minipage}
} \par}

\paragraph{Example summary:} The Persuader has employed a credibility appeal mentioning the accomplishments of the charity. The Persuadee seems to respond positively and wants to learn more. 

{\centering
\fbox{
\begin{minipage}{\linewidth}
\textbf{Goal proposal prompt:} \\
The following is some background information about Save the Children.
Save the Children is head-quartered in London, and they work to help fight poverty around the world.
Children need help in developing countries and war zones. Small donations like \$1 or \$2 go a long way to help. \\

The following is a conversation between a Persuader and a Persuadee about a charity called Save the Children. The Persuader is trying to persuade the Persuadee to donate to Save the Children. \\
\textcolor{blue}{[DIALOGUE]}\\
Please propose a future \{positive|negative\} goal. The goal should include how the Persuadee responds with an assessment of how likely they are to donate.\\

Here are examples below: \\
\textcolor{blue}{[DIALOGE+THOUGHT+GOAL]}
\end{minipage}
} \par} 

\paragraph{Example negative goal:} The Persuadee distrusts the Persuader and feels they are simply employing a tactic and not speaking genuinely.

{\centering
\fbox{
\begin{minipage}{\linewidth}
\textbf{Refinement prompt:} \\
The following is some background information about Save the Children.
Save the Children is head-quartered in London, and they work to help fight poverty around the world.
Children need help in developing countries and war zones. Small donations like \$1 or \$2 go a long way to help. \\

The following is a conversation between a Persuader and a Persuadee about a charity called Save the Children. The Persuader is trying to persuade the Persuadee to donate to Save the Children. \\
\textcolor{blue}{[DIALOGUE]}\\
Here is the Persuader's current strategy on how to proceed, as well as an assessment of possible positive and negative outcomes of this strategy made by a critic trained on lots of data: \\
\textcolor{blue}{[THOUGHT+NATURAL LANGUAGE VALUE]} \\
First, determine whether or not the current pro-
posed strategy will lead to success. If not, in a few
sentences, diagnose a possible reason for failure
and devise a new strategy that aims to mitigate the
same failure,
\end{minipage}
} \par} 

\paragraph{Example original thought:} As the Persuadee does not know about the charity, I should talk about how small donations can lead to a large impact.

\paragraph{Example refinement:} I should highlight the credentials of the charity including things that it has accomplished with the donations so far.      

\subsection{Persuasion Training Details}

\begin{table}[H]
\centering
\begin{tabular}{lrr}
\toprule
Hyperparameter & \multicolumn{2}{r}{Setting } \\
\midrule
\midrule
Hidden-layer size && 128*128 \\
IQL $\tau$ && 0.8 \\
Discount factor && 0.99 \\
Batch size && 32 \\
Target network update $\alpha$ && 0.005  \\
Number of updates per iteration && 50 \\
Number of iterations && 100 \\
Optimizer && AdamW \\
Learning rate && 1e-4 \\

\bottomrule
\end{tabular}
\end{table}

\end{document}